\documentclass[conference]{IEEEtran}
\IEEEoverridecommandlockouts
\usepackage{cite}
\usepackage{amsmath,amssymb,amsfonts}
\usepackage{algorithmic}
\usepackage{graphicx}
\usepackage{textcomp}
\usepackage{xcolor}
\usepackage{enumitem}
\usepackage{makecell}
\def\BibTeX{{\rm B\kern-.05em{\sc i\kern-.025em b}\kern-.08em
    T\kern-.1667em\lower.7ex\hbox{E}\kern-.125emX}}
\begin{document}

\title{Privacy-Preserving Semantic Segmentation without Key Management\\
\thanks{This work was supported in part by JSPS KAKENHI Grant Number 25K07750.}
}

\author{\IEEEauthorblockN{Mare Hirose}
\IEEEauthorblockA{\textit{Chiba University} \\
Chiba, Japan \\
marehirose@chiba-u.jp}
\and
\IEEEauthorblockN{Shoko Imaizumi}
\IEEEauthorblockA{\textit{Chiba University} \\
Chiba, Japan \\
imaizumi@chiba-u.jp}
\and
\IEEEauthorblockN{Hitoshi Kiya}
\IEEEauthorblockA{\textit{Tokyo Metropolitan University} \\
Tokyo, Japan \\
kiya@tmu.ac.jp}
}

\maketitle

\begin{abstract}
This paper proposes a novel privacy-preserving semantic segmentation method  that can use independent keys for each client and image. In the proposed method, the model creator and each client encrypt images using locally generated keys, and model training and inference are conducted on the encrypted images. To mitigate performance degradation, an image encryption method is applied to model training in addition to the generation of test images. In experiments, the effectiveness of the proposed method is confirmed on the Cityscapes dataset under the use of a vision transformer-based model, called SETR.
\end{abstract}
\vspace{-5pt}

\vspace{-3.5pt}
\section{Introduction}
Recently, deep learning has been increasingly adopted across a wide range of applications, and model training and segmentation inference are often conducted in cloud environments. However, cloud environments raise privacy concerns because sensitive information in images may be exposed to untrusted parties \cite{cloud, cloud2}. To address this issue, many studies have explored novel methods for privacy-preserving learning by encrypting input data \cite{cloud2, enc}.

Methods with perceptual encryption were demonstrated to be effective in terms of lightweight computation and segmentation accuracy. However, previous segmentation methods with perceptual encryption must share keys between the model creator and the clients \cite{seg1, seg2}. This requirement means that both the model creator and clients are trusted and maintain strict key management. Thus, if the key is compromised, the impact is extended to all clients and images using the same key.

Accordingly, we propose a novel semantic segmentation method for the first time that enables independent key encryption for each client and each image to strengthen privacy protection. In addition, the model creator does not need to share keys with clients. In experiments, we evaluate the proposed method using a Segmentation Transformer (SETR) \cite{SETR} in terms of model performance and privacy protection.

\section{Proposed method}
\subsection{Overview of proposed method}
\begin{figure}[t]
  \centering
  \includegraphics[scale=0.28]{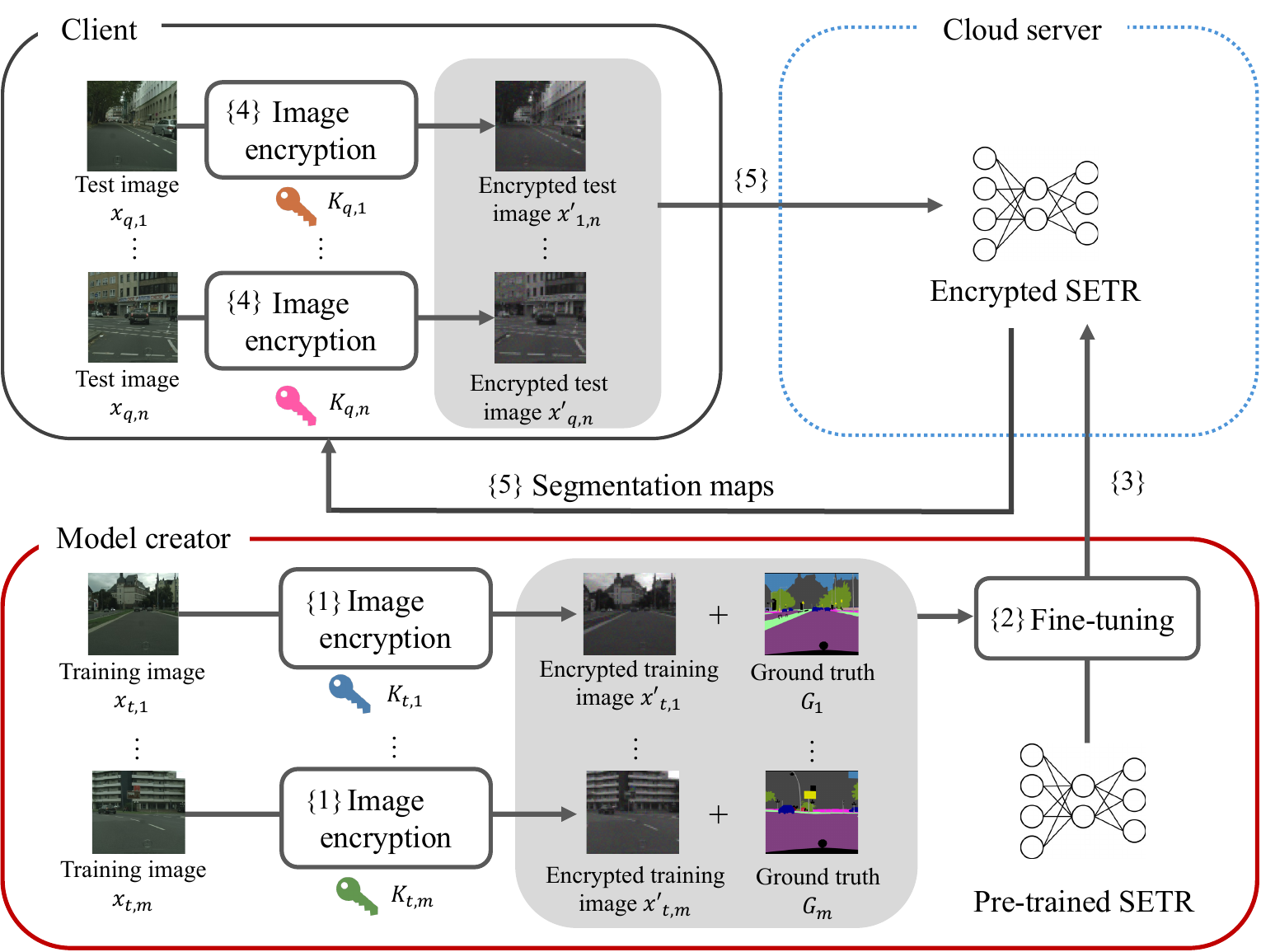}
  \caption{Overview of proposed method.}
  \label{fig:overview}
\end{figure}

Fig. \ref{fig:overview} illustrates an overview of the proposed method. A model creator encrypts the training images using independent encryption keys prepared for each image (\{1\} in Fig. \ref{fig:overview}).
Then, the creator fine-tunes a pre-trained model on the encrypted images and their ground truth (\{2\}) and uploads the resulting encrypted model to a cloud environment (\{3\}).
At test time, on the client side, the client encrypts each test image with independent encryption keys prepared for each image (\{4\}). Finally, it sends the encrypted test images to the encrypted model and receives the inference results (\{5\}).

As shown in Fig. \ref{fig:overview}, this method can change encryption keys for each image. Therefore, even if an encryption key is compromised or misused, the method reduces the risk of privacy violations.

\subsection{Image encryption}
Image encryption used in the proposed method is carried out in accordance with the embedding structure of ViT-based SETR models.

\begin{enumerate}[label=Step \arabic*:, leftmargin=*]  
  \item Divide an image into non-overlapped blocks with a size of $M \times M$, which is the same size as the patch size of SETR.
  \item Divide each block into non-overlapped sub-blocks with a size of $M_s \times M_s$.
  \item Shuffle pixels within each sub-block for each color channel and then permute the order of the color channels.
\end{enumerate}

The proposed method can apply independent keys to each block and sub-block. This feature enhances the robustness against image restoration attacks. Unlike previous methods, different keys are applied to each client and image, so the method does not need key management.

\section{Experiments}

\begin{table}[t]
    \centering
    \caption{Hyperparameters for training.}
    \label{tab:hyper}
    \renewcommand{\arraystretch}{1.1}
    \begin{tabular}{lc}
    \Xhline{0.8pt}
     Backbone & vit\_base\_patch16\_384
  \\
     Decoder & MLA\\
     Optimizer & SGD\\
     Learning rate & 1e-3\\
     Momentum & 0.9\\
     Weight decay & 5e-4\\
     Scheduler & Polynomial decay\\
     \# of iterations & 89,250\\
    \Xhline{0.8pt}
    \end{tabular}
    
\end{table}

\begin{figure}[t]
  \centering
  \includegraphics[scale=0.33]{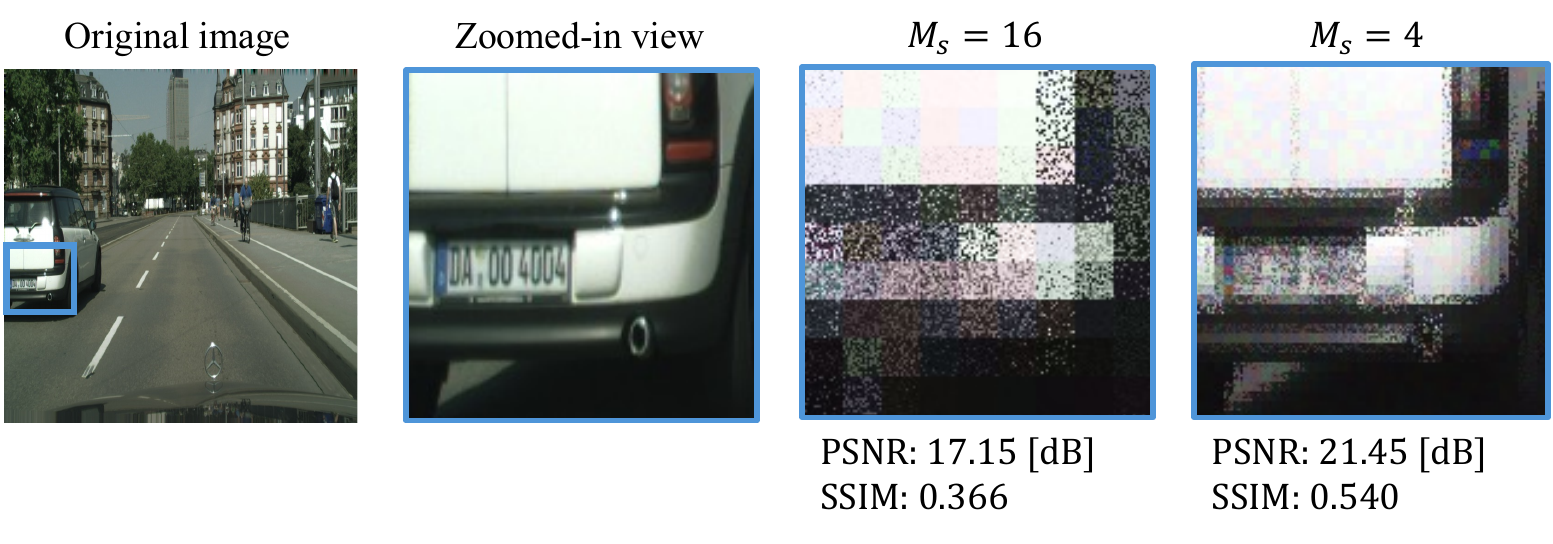}
  \vspace{-10pt}
  \caption{Examples of encrypted images.}
  \label{fig:enc}
\end{figure}


\begin{table}[t]
    \centering
    \caption{Accuracy of estimated semantic segmentation maps.}
    \label{tab:acc}
    \renewcommand{\arraystretch}{1.1}
    \begin{tabular}{lccc}
    \Xhline{0.8pt}
    Method & aAcc & mIoU & mAcc\\
    \hline
     Baseline & 92.90 & 61.63 & 71.73 \\
     $M_s$ = 16 & 88.44 & 44.51 & 53.23 \\
     $M_s$ = 8 & 90.26 & 50.00 & 59.42 \\
     $M_s$ = 4 & 91.65 & 56.13 & 68.18 \\
    \Xhline{0.8pt}
    \end{tabular}
\end{table}

\begin{figure}[t]
  \centering
  \includegraphics[scale=0.4]{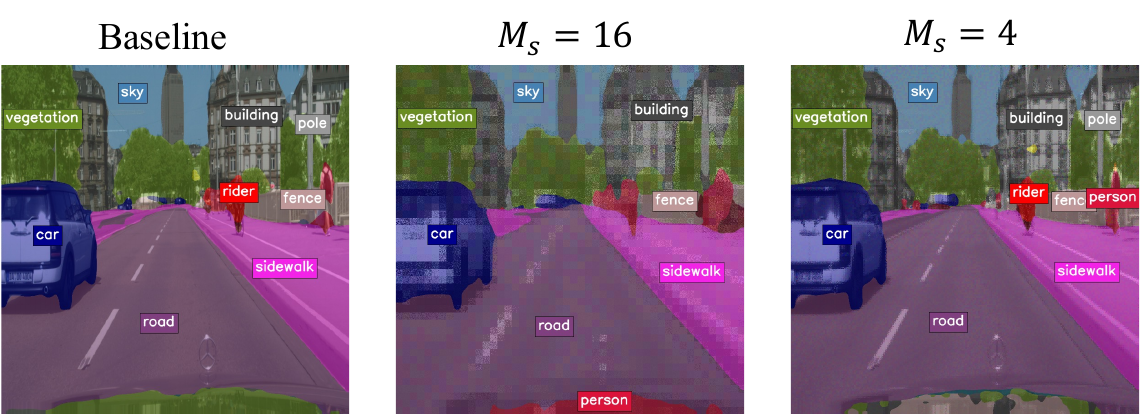}
  \caption{Examples of segmentation maps.}
  \label{fig:segmap}
\end{figure}

\subsection{Setup}
We conducted semantic-segmentation experiments on the Cityscapes dataset. The Cityscapes dataset consists of 2,975 training images, 500 validation images, and 1,525 test images. In the experiments, we used training images for model training and validation images for inference. All images were resized to $768 \times 768$ pixels. 
For the encryption, we set the block size $M$ to 16 and the sub-block size $M_s$ to 16, 8, and 4.
We used SETR from OpenMMLab as the segmentation model. Table \ref{tab:hyper} lists the hyperparameters used for training.

Average accuracy (aAcc), mean Intersection over Union (mIoU) \cite{mIoU}, and mean accuracy (mAcc) were used as the evaluation metrics of model performance. Each metric takes a value between 0 and 100, where higher values indicate better model performance.

\subsection{Results}
Fig. \ref{fig:enc} shows examples of encrypted images. As the sub-block size increased, the overall structure of the original image became more effectively obfuscated, but sensitive information such as car plate numbers could be protected even when $M_s = 4$ was selected.

Table \ref{tab:acc} shows the model performance, where ``Baseline'' denotes the results obtained by training and evaluating the model on plain images. The proposed method consistently preserved aAcc for all $M_s$, even with independent key encryption.
Meanwhile, mAcc and mIoU exhibited a noticeable degradation under encryption. These results indicate that reducing $M_s$ leads to model performance closer to ``Baseline.'' Furthermore, as shown in Fig. \ref{fig:segmap}, the overall quality of the segmentation maps improved as the sub-block size $M_s$ decreased. In particular, a smaller $M_s$ resulted in sharper object boundaries.

Overall, the proposed method suppressed the accuracy drop even with independent key encryption, while the sub-block size affected the trade-off between model performance and privacy protection.

\section{Conclusions}
We proposed a semantic segmentation method with independent keys applied to each client and each image. We demonstrated that the semantic segmentation can be performed under independent key encryption. Our results further reveal a trade-off between model performance and privacy protection. Future work will focus on developing training methods that improve model performance to the level achieved with training and inference on plain images.


\begin{thebibliography}{00}
\bibitem{cloud} Y. Liu, H. Chen, and Z. Yang, ``Enforcing End-to-end Security for Remote Conference Applications,'' in \textit{Proc. IEEE Symp. Secur. Priv.}, San Francisco,
CA, USA, 2024, pp. 2630--2647.
\bibitem{cloud2} K. Madono, M. Tanaka, M. Onishi, and T. Ogawa, ``Block-wise Scrambled Image Recognition Using Adaptation Network,'' 
in \textit{Proc. Workshop on
Artif. Intell. Things (AAAI-WS)}, New York, NY, USA, 2020.
\bibitem{enc} M. Hirose, S. Imaizumi, and H. Kiya, ``Learnable Image Encryption Without Key Management for Privacy-Preserving Vision Transformer,'' \textit{IEEE Access}, vol. 13, pp. 201351--201362, 2025.
\bibitem{seg1} H. Sueyoshi, K. Nishikawa and H. Kiya, ``A Privacy-Preserving Semantic-Segmentation Method Using Domain-Adaptation Technique,'' in \textit{Proc. IEEE GCCE}, Osaka, Japan, 2025, pp. 37--40.
\bibitem{seg2} H. Kiya, T. Nagamori, S. Imaizumi, S. Shiota, ``Privacy-Preserving Semantic Segmentation Using Vision Transformer,'' \textit{J. Imaging}, vol. 8, no. 9, 2022.
\bibitem{SETR} S. Zheng et al., ``Rethinking Semantic Segmentation from a Sequence-to-Sequence Perspective with Transformers,'' in \textit{Proc. CVPR}, 2021, pp. 6881--6890.
\bibitem{mIoU} M. Everingham, L. Van Gool, C. K. I. Williams, J. Winn, and A. Zisserman, ``The PASCAL Visual Object Classes (VOC) Challenge,'' \textit{Int. J. Comput. Vis.}, vol. 88, pp. 303--338, 2010.



\end{thebibliography}
\end{document}